\pdfoutput=1

\documentclass[11pt]{article}

\usepackage[]{emnlp2021}
\usepackage{times}
\usepackage{latexsym}
\usepackage{hyperref}

\usepackage[T1]{fontenc}

\usepackage[utf8]{inputenc}

\usepackage{microtype}

\usepackage{bm}
\usepackage{amsmath}
\usepackage{url}
\usepackage{graphicx}
\usepackage{float} 
\usepackage{subfigure}
\usepackage{multirow}
\usepackage{makecell}
\usepackage{xpinyin}
\usepackage{enumerate}

\usepackage{xcolor}
\usepackage{soul}
\newcolumntype{L}[1]{>{\raggedright\let\newline\\\arraybackslash\hspace{0pt}}m{#1}}
\setul{2pt}{2pt}
\usepackage{booktabs}
\usepackage{amsmath,amsfonts}
\usepackage[utf8]{inputenc}
\usepackage{cleveref}

\crefname{section}{§}{§§}
\Crefname{section}{§}{§§}
\usepackage{enumitem}
\setenumerate[1]{itemsep=0pt,partopsep=0pt,parsep=\parskip,topsep=5pt}
\setitemize[1]{itemsep=0pt,partopsep=0pt,parsep=\parskip,topsep=5pt}
\setdescription{itemsep=0pt,partopsep=0pt,parsep=\parskip,topsep=5pt}
\usepackage{arydshln}

\usepackage[linesnumbered,ruled,vlined]{algorithm2e}
\title{Multi-Task Learning for Neural Chat Translation}
\title{Towards Making the Most of Dialogue Characteristics \\for Neural Chat Translation}

\author{
  Yunlong Liang\textsuperscript{1}\thanks{ \ \ Equal contribution. Work was done when Liang and Zhou were interning at Pattern Recognition Center, WeChat AI, Tencent Inc, China.}, 
  Chulun Zhou\textsuperscript{2}\footnotemark[1],
  Fandong Meng\textsuperscript{3}, 
  \textbf{Jinan Xu}\textsuperscript{1}\thanks{ \ \ Jinan Xu is the corresponding author.}, \\
  \textbf{Yufeng Chen}\textsuperscript{1}, 
  \textbf{Jinsong Su}\textsuperscript{2}
  and \textbf{Jie Zhou}\textsuperscript{3}\\
  \textsuperscript{1}Beijing Key Lab of Traffic Data Analysis and Mining, Beijing Jiaotong University \\
  \textsuperscript{2}School of Informatics, Xiamen University \\
  \textsuperscript{3}Pattern Recognition Center, WeChat AI, Tencent Inc, China \\
  \texttt{\{yunlongliang,chenyf,jaxu\}@bjtu.edu.cn} \ \ \ \  \texttt{clzhou@stu.xmu.edu.cn}\\
  \texttt{jssu@xmu.edu.cn} \ \ \ \  \texttt{\{fandongmeng,withtomzhou\}@tencent.com} \\
}

\begin{document}
\maketitle
\begin{abstract}
Neural Chat Translation (NCT) aims to translate conversational text between speakers of different languages. Despite the promising performance of sentence-level and context-aware neural machine translation models, there still remain limitations in current NCT models because the inherent dialogue characteristics of chat, such as dialogue coherence and speaker personality, are neglected. In this paper, we propose to promote the chat translation by introducing the modeling of dialogue characteristics into the NCT model. To this end, we design four auxiliary tasks including \emph{monolingual response generation}, \emph{cross-lingual response generation}, \emph{next utterance discrimination}, and \emph{speaker identification}. Together with the main chat translation task, we optimize the NCT model through the training objectives of all these tasks. By this means, the NCT model can be enhanced by capturing the inherent dialogue characteristics, thus generating more coherent and speaker-relevant translations. Comprehensive experiments on four language directions (English$\Leftrightarrow$German and English$\Leftrightarrow$Chinese) verify the effectiveness and superiority of the proposed approach.
\end{abstract}

\section{Introduction}
A cross-lingual conversation involves participants that speak in different languages (\emph{e.g.}, one speaking in English and another in Chinese as shown in \autoref{fig:ctx_case}), where a chat translator can be applied to help participants communicate in their individual native languages. The chat translator converts the language of bilingual conversational text in both directions, \emph{e.g.} from English to Chinese and vice versa \cite{farajian-etal-2020-findings}. With more international communication worldwide, the chat translation task becomes more important and has a wider range of applications. 
\textbf{\begin{figure}[t]
    \centering
    \includegraphics[width=0.49\textwidth]{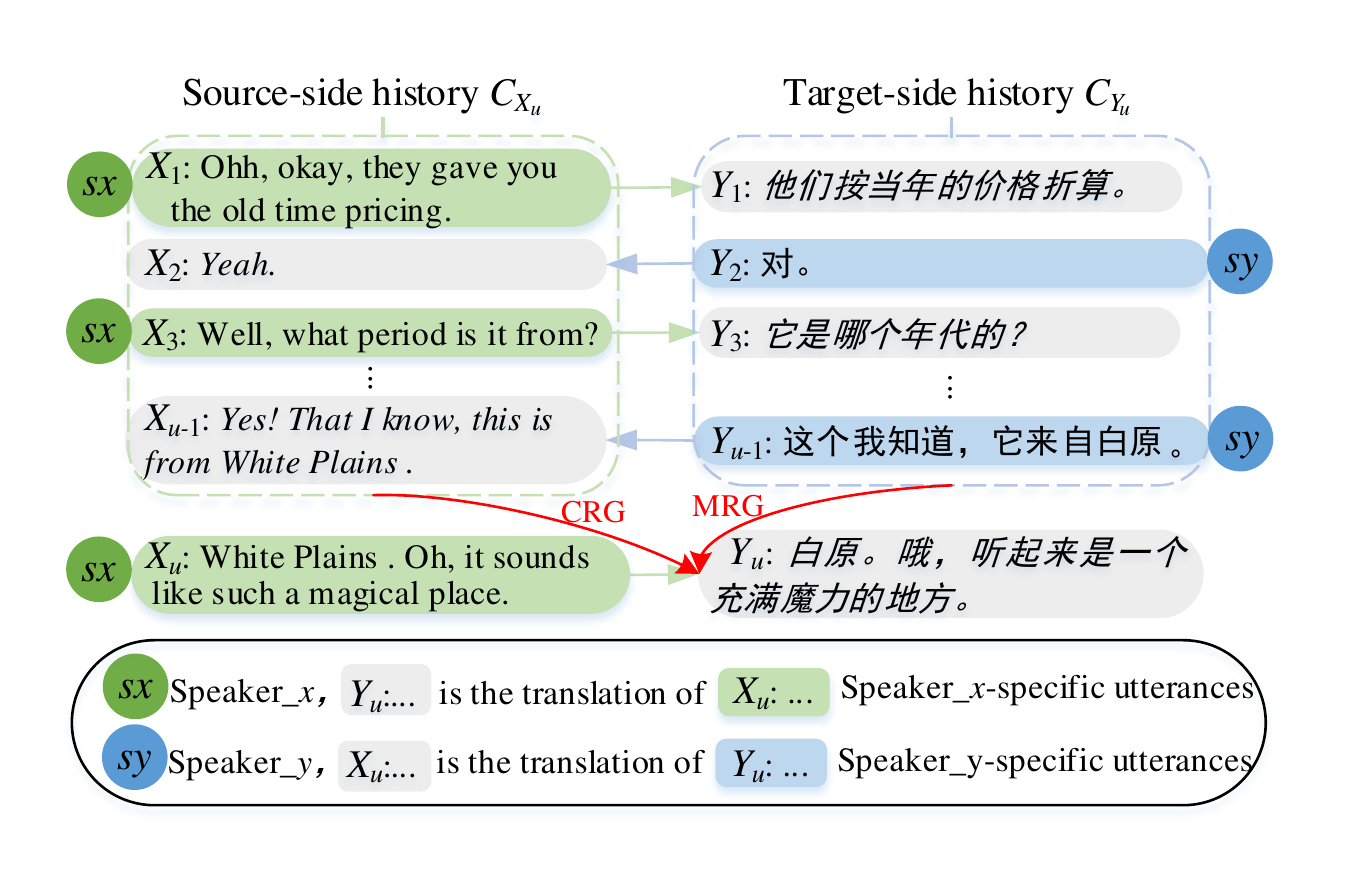}
    \caption{A dialogue example (En$\Leftrightarrow$Zh) when translating the utterance $X_{u}$. CRG: cross-lingual response generation. MRG: monolingual response generation.
    }
    \label{fig:ctx_case}
\end{figure}}

In recent years, although sentence-level Neural Machine Translation (NMT) models~\cite{NIPS2014_a14ac55a,vaswani2017attention,hassan2018achieving,meng2019dtmt,yanetal2020multi,zhangetal2019bridging} have achieved remarkable progress and can be directly used as the chat translator, they often lead to incoherent and speaker-irrelevant translations~\cite{mirkin-etal-2015-motivating,wang-etal-2017-semantics,laubli-etal-2018-machine,toral-etal-2018-attaining} due to ignoring the chat history that contains useful contextual information. To exploit chat history, context-aware NMT models (\citealp{tiedemann-scherrer-2017-neural,maruf-haffari-2018-document,bawden-etal-2018-evaluating,miculicich-etal-2018-document,tu-etal-2018-learning,voita-etal-2018-context,voita-etal-2019-context,voita-etal-2019-good,wang-etal-2019-one,maruf-etal-2019-selective,DBLP:journals/corr/abs-2006-04721,ma-etal-2020-simple}, etc) can also be directly adapted to chat translation. However, their performances are usually limited because of lacking the modeling of the inherent dialogue characteristics (\emph{e.g.}, the dialogue coherence and speaker personality), which matter for chat translation task as pointed out by~\citet{farajian-etal-2020-findings}.

In this paper, we propose a \textbf{C}oherence-\textbf{S}peaker-\textbf{A}ware NCT (CSA-NCT) training framework to improve the NCT model by making use of dialogue characteristics in conversations. Concretely, from the perspectives of dialogue coherence and speaker personality, we design four auxiliary tasks along with the main chat translation task. For dialogue coherence, there are three tasks (two generation tasks and one discrimination task), namely \emph{monolingual response generation}, \emph{cross-lingual response generation}, and \emph{next utterance discrimination}. Specifically, as shown in~\autoref{fig:ctx_case}, (1) the monolingual response generation task aims to generate the coherent corresponding utterance in target language given the dialogue history context of the same language. Similarly, (2) the cross-lingual response generation task is to leverage the dialogue history context in source language to generate the coherent corresponding utterance in target language. Besides the above two generation tasks, (3) the next utterance discrimination task focuses on distinguishing whether the translated text is coherent to be the next utterance of the given dialogue history context. Moreover, for speaker personality, (4) we design the speaker identification task that judges whether the translated text is consistent with the personality of its original speaker. Together with the main chat translation task, the NCT model is optimized through the joint objectives of all these auxiliary tasks. In this way, the model is enhanced to capture dialogue coherence and speaker personality in conversation, which thus can generate more coherent and speaker-relevant translations.

We validate our CSA-NCT framework on the datasets of different language pairs: BConTrasT~\cite{farajian-etal-2020-findings} (En$\Leftrightarrow$De\footnote{En$\Leftrightarrow$De:\,English$\Leftrightarrow$German.}) and BMELD~\cite{liang-etal-2021-modeling} (En$\Leftrightarrow$Zh\footnote{En$\Leftrightarrow$Zh:\,English$\Leftrightarrow$Chinese.}). The experimental results show that our model achieves consistent improvements on four translation tasks in terms of both BLEU~\cite{papineni2002bleu} and TER~\cite{snover2006study}, demonstrating its effectiveness and generalizability. Human evaluation further suggests that our model can generate more coherent and speaker-relevant translations compared to the existing related methods. 

Our contributions are summarized as follows:
\begin{itemize}
\item To the best of our knowledge, we are the first to incorporate the dialogue coherence and speaker personality into neural chat translation.

\item We propose a multi-task learning framework with four auxiliary tasks to help the NCT model generate more coherent and speaker-relevant translations.

\item Extensive experiments on datasets of different language pairs demonstrate that our model with multi-task learning achieves the state-of-the-art performances on the chat translation task and significantly outperforms the existing sentence-level/context-aware NMT models.\footnote{The code is publicly available at: \url{https://github.com/XL2248/CSA-NCT}}
\end{itemize}

\section{Background}

\paragraph{Sentence-Level NMT.}
\label{sec:length}
Given an input sequence $X$$=$$\{x_i\}_{i=1}^{|X|}$, the goal of the sentence-level NMT model is to generate its translation $Y$$=$$\{y_i\}_{i=1}^{|Y|}$. The model is optimized through the following objective:
\begin{equation}
\label{eq:nmt}
    \mathcal{L}_{\text{S-NMT}} = -\sum_{t=1}^{|Y|}\mathrm{log}(p(y_t|X, y_{<t})).
\end{equation}

\paragraph{Context-Aware NMT.}
As in~\cite{ma-etal-2020-simple}, given a paragraph of input sentences $D^X=\{X_j\}_{j=1}^{J}$ in source language and its corresponding translations $D^Y=\{Y_j\}_{j=1}^{J}$ in target language with $J$ paired sentences, the training objective of a context-aware NMT model can be formalized as
\begin{equation}
\label{eq:dnmt}
    \mathcal{L}_{\text{C-NMT}} = -\sum_{j=1}^{J}\mathrm{log}(p(Y_j|X_j, X_{<j}, Y_{<j})),
\end{equation}
where $X_{<j}$ and $Y_{<j}$ are the preceding contexts of the $j$-th input source sentence and the $j$-th target translation, respectively. 

\section{CSA-NCT Training Framework}
In this section, we introduce the proposed CSA-NCT training framework, which aims to improve the NCT model with four elaborately designed auxiliary tasks. In the following subsections, we first describe the problem formalization (\autoref{sec:pf}) and the NCT model (\autoref{sec:nct}). Then, we introduce each auxiliary task in detail (\autoref{sec:aux}). Finally, we elaborate the process of training and inference (\autoref{sec:ti}).
\subsection{Problem Formalization}
\label{sec:pf}
In the scenario of this paper, the chat involves two speakers ($sx$ and $sy$) speaking in two languages. As shown in~\autoref{fig:ctx_case}, we assume the two speakers have alternately given utterances in their individual languages for $u$ turns, resulting in $X_1, X_2, X_3, X_4, X_5,..., X_{u-1}, X_u$ and $Y_1, Y_2, Y_3, Y_4, Y_5,...,Y_{u-1}, Y_u$ on the source and target sides, respectively. Among these utterances, $X_1, X_3, X_5,..., X_u$ are originally spoken by the speaker $sx$ and $Y_1, Y_3, Y_5,..., Y_u$ are the corresponding translations in target language. Analogously, $Y_2, Y_4, Y_6,..., Y_{u-1}$ are originally spoken by the speaker $sy$ and $X_2, X_4, X_6,..., X_{u-1}$ are the translated utterances in source language. 

According to languages, we define the dialogue history context of $X_u$ on the source side as $\mathcal{C}_{X_u}$=\{$X_1, X_2, X_3, X_4, X_5,..., X_{u-1}\}$ and that of $Y_u$ on the target side as $C_{Y_u}$=\{$Y_1, Y_2, Y_3, Y_4, Y_5,..., Y_{u-1}\}$. According to original speakers, on the target side, we define the speaker $sx$-specific dialogue history context of $Y_u$ as the partial set of its preceding utterances $\mathcal{C}_{Y_u}^{sx}$=\{$Y_1, Y_3, Y_5,..., Y_{u-2}\}$ and the speaker $sy$-specific dialogue history context of $Y_u$ as $\mathcal{C}_{Y_u}^{sy}$=\{$Y_2, Y_4, Y_6,..., Y_{u-1}\}$.\footnote{For each item of \{$C_{X_u}$, $C_{Y_u}$, $C_{Y_u}^{sx}$, $C_{Y_u}^{sy}$\}, taking $C_{X_u}$ for instance, we add the special token `[cls]' tag at the head of it and use another special token `[sep]' to delimit its included utterances, as in~\cite{bert}. }

Based on the above formulations, the goal of an NCT model is to translate $X_u$ to $Y_u$ with certain types of dialogue history context.\footnote{Here, we just take one translation direction (\emph{i.e.}, En$\Rightarrow$Zh) as an example, which is similar for other directions.} Next, we will descibe the NCT model in our CSA-NCT training framework.

\subsection{The NCT Model}
\label{sec:nct}
The NCT model is based on transformer~\cite{vaswani2017attention}, which is composed of an encoder and a decoder as shown in \autoref{fig:model}.

\paragraph{Encoder.}
\label{sec:enc}
Following~\cite{ma-etal-2020-simple}, the encoder takes $[\mathcal{C}_{X_u}$; $X_u]$ as input, where $[;]$ denotes the concatenation. In addition to the conventional embedding layer with only word embedding $\mathbf{WE}$ and position embedding $\mathbf{PE}$, we additionally add a speaker embedding $\mathbf{SE}$ and a turn embedding $\mathbf{TE}$. The final embedding $\mathbf{B}(x_i)$ of the input word $x_i$ can be written as
\begin{equation}\label{input_embed}\nonumber
\mathbf{B}(x_i) = \mathbf{WE}({x_i}) + \mathbf{PE}({x_i}) + \mathbf{SE}({x_i}) + \mathbf{TE}({x_i}),
\end{equation}
where $\mathbf{WE}\in\mathbb{R}^{|V|\times{d}}$, $\mathbf{SE}\in\mathbb{R}^{2\times{d}}$ and $\mathbf{TE}\in\mathbb{R}^{|T|\times{d}}$.\footnote{$|V|$, $|T|$ and $d$ denote the size of shared vocabulary, maximum dialogue turns, and the hidden size, respectively.}

Then, the embedding is fed into the NCT encoder that has $L$ identical layers, each of which is composed of a self-attention ($\mathrm{SelfAtt}$) sub-layer and a feed-forward network ($\mathrm{FFN}$) sub-layer.\footnote{The layer normalization is omitted for simplicity.} Let $\mathbf{h}^{l}_e$ denote the hidden states of the $l$-th encoder layer, it is calculated as the following equations:
\begin{equation}
\begin{split}
    \mathbf{z}^l_e &= \mathrm{SelfAtt}(\mathbf{h}^{l-1}_e) + \mathbf{h}^{l-1}_e,\,\nonumber\\
    \mathbf{h}^l_e &= \mathrm{FFN}(\mathbf{z}^l_e) + \mathbf{z}^l_e,\ \nonumber
\end{split}
\end{equation}
where $\mathbf{h}^{0}_e$ is initialized as the embedding of input words.
Particularly, words in $\mathcal{C}_{X_u}$ can only be attended to by those in $X_u$ at the first encoder layer while $\mathcal{C}_{X_u}$ is masked at the other layers, which is the same implementation as in~\cite{ma-etal-2020-simple}.
\textbf{\begin{figure}[t]
    \centering
    \includegraphics[width=0.49\textwidth]{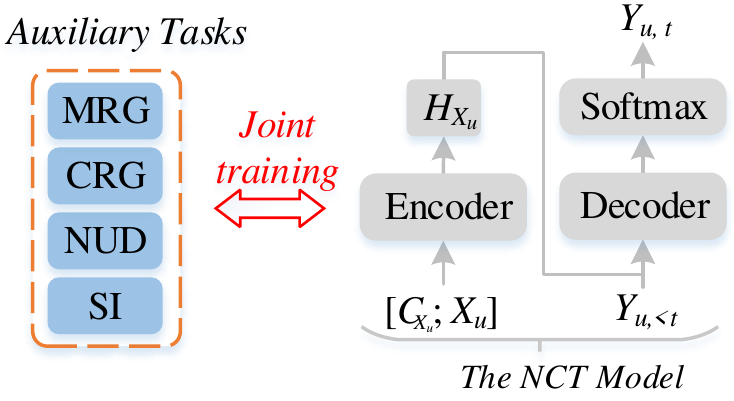}
    \caption{Architecture of the proposed CSA-NCT framework. The right part is the general NCT model, which is enhanced by four auxiliary tasks. The four auxiliary tasks including \emph{monolingual response generation} (MRG), \emph{cross-lingual response generation} (CRG), \emph{next utterance discrimination} (NUD), and \emph{speaker identification} (SI), are proposed to improve the coherence and speaker relevance of chat translation, which are presented in~\autoref{fig:subtask} in detail. 
    }
    \label{fig:model}
\end{figure}}

\begin{figure*}[ht]
\centering
  \includegraphics[width = 0.98\textwidth]{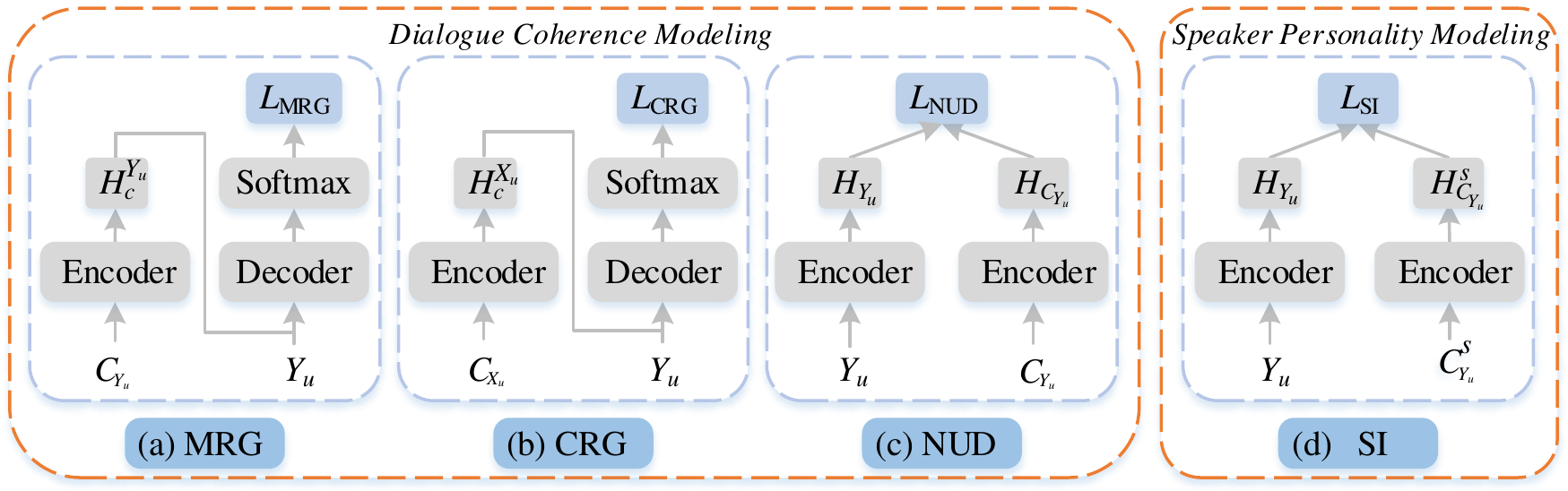}
\caption{Overview of four auxiliary tasks. The encoder and the decoder of auxiliary tasks are shared with the NCT model. The encoder encodes not only source-side but also target-side history context to enhance its ability of representation. } 
\label{fig:subtask}
\end{figure*}

\paragraph{Decoder.}
\label{sec:dec}
The decoder also consists of $L$ identical layers, each of which additionally includes a cross-attention ($\mathrm{CrossAtt}$) sub-layer compared to the encoder. Let $\mathbf{h}^{l}_d$ denote the hidden states of the $l$-th decoder layer, it is computed as
\begin{equation}
\begin{split}
\label{eq:trans_de}
    \mathbf{z}^l_d &= \mathrm{SelfAtt}(\mathbf{h}^{l-1}_d) + \mathbf{h}^{l-1}_d,\ \nonumber\\
    \mathbf{c}^l_d &= \mathrm{CrossAtt}(\mathbf{z}^{l}_d, \mathbf{h}_e^{L}) + \mathbf{z}^l_d,\ \nonumber\\
    \mathbf{h}^l_d &= \mathrm{FFN}(\mathbf{c}^l_d) + \mathbf{c}^l_d,\ \nonumber
\end{split}
\end{equation}
where $\mathbf{h}^L_e$ is the top-layer encoder hidden states.

At each decoding time step $t$, $\mathbf{h}^{L}_{d,t}$ is fed into a linear transformation layer and a softmax layer to predict the probability distribution of the next target token:
\begin{equation}
\resizebox{1.05\hsize}{!}{$
\begin{split}
    p(Y_{u,t}|Y_{u,<t}, X_u, \mathcal{C}_{X_u}) &= \mathrm{Softmax}(\mathbf{W}_o\mathbf{h}^{L}_{d,t}+\mathbf{b}_o),\nonumber
\end{split}
$}
\end{equation}
where $Y_{u,<t}$ denotes the preceding tokens before the $t$-th time step in the utterance $Y_u$, $\mathbf{W}_o \in \mathbb{R}^{|V|\times d}$ and $\mathbf{b}_o \in \mathbb{R}^{|V|}$ are trainable parameters. 

Finally, the training objective is as follows:
\begin{equation}
\begin{split} 
\label{eq:cnmt}
    \mathcal{L}_{\text{NCT}} = -\sum_{t=1}^{|Y_u|}\mathrm{log}(p(Y_{u,t}|Y_{u,<t}, X_u, \mathcal{C}_{X_u})).
\end{split}
\end{equation}

\subsection{Auxiliary Tasks}
\label{sec:aux}
We elaborately design four auxiliary tasks to incorporate the modeling of dialogue characteristics. The four auxiliary tasks are divided into two groups. 
The first group is for dialogue coherence modeling while the second is for speaker personality modeling. Together with the main chat translation task, the NCT model can be enhanced to generate more coherent and speaker-relevant translations through multi-task learning.

\subsubsection{Dialogue Coherence Modeling} 
Many studies~\cite{articleKuang,wang-etal-2019-answer,Xiong_He_Wu_Wang_2019,ijcai2019-727,huang-etal-2020-grade} have indicated that the modeling of global textual coherence can lead to more coherent text generation. Inspired by this, we add two response generation tasks and an utterance discrimination task during the NCT model training. All the three tasks are related to the dialogue coherence of conversations, thus introducing the modeling of dialogue coherence into the NCT model.

\paragraph{Monolingual Response Generation (MRG).} As illustrated in \autoref{fig:subtask}(a), given the dialogue history context $\mathcal{C}_{Y_u}$ in target language, the MRG task forces the NCT model to generate the corresponding utterance $Y_u$ coherent to $\mathcal{C}_{Y_u}$. Particularly, we first use the encoder of the NCT model to encode $\mathcal{C}_{Y_u}$, and then use the NCT decoder to predict $Y_u$. The training objective of this task can be formulated as:
\begin{equation}
\begin{split}
    \mathcal{L}_{\text{MRG}}  =-\sum^{|Y_u|}_{t=1}\mathrm{log}(p(Y_{u,t}|\mathcal{C}_{Y_u},Y_{u,<t})),\\
    {p}(Y_{u,t}|\mathcal{C}_{Y_u},Y_{u,<t}) = \mathrm{Softmax}(\mathbf{W}_{m}\mathbf{h}^{L}_{d,t}+\mathbf{b}_{m}),\nonumber
\end{split}
\end{equation}
where $\mathbf{h}^{L}_{d,t}$ is the top-layer decoder hidden state at the $t$-th decoding step, $\mathbf{W}_{m}$ and $\mathbf{b}_{m}$ are trainable parameters. 

\paragraph{Cross-lingual Response Generation (CRG).} 
The CRG task is similar to the MRG as shown in \autoref{fig:subtask}(b), where the NCT model is trained to generate the corresponding utterance $Y_u$ in target language which is coherent to the given dialogue history context $\mathcal{C}_{X_u}$ in source language. We first use the encoder of the NCT model to encode $\mathcal{C}_{X_u}$, and then use the NCT decoder to predict $Y_u$. The training objective of this task can be formulated as:
\begin{equation}
\begin{split}
    \mathcal{L}_{\text{CRG}}  =-\sum^{|Y_u|}_{t=1}\mathrm{log}(p(Y_{u,t}|\mathcal{C}_{X_u},Y_{u,<t})),\\
    {p}(Y_{u,t}|\mathcal{C}_{X_u},Y_{u,<t}) = \mathrm{Softmax}(\mathbf{W}_{c}\mathbf{h}^{L}_{d,t}+\mathbf{b}_{c}),\nonumber
\end{split}
\end{equation}
where $\mathbf{h}^{L}_{d,t}$ denotes the top-layer decoder hidden state at the $t$-th decoding step, $\mathbf{W}_{crg}$ and $\mathbf{b}_{crg}$ are trainable parameters. 

Note that in the above two response generation tasks, we use the same set of NCT model parameters except for the softmax layer (\emph{i.e.}, $\mathbf{W}_{m}$, $\mathbf{b}_{m}$, $\mathbf{W}_{c}$ and $\mathbf{b}_{c}$).

\paragraph{Next Utterance Discrimination (NUD).} 
As shown in \autoref{fig:subtask}(c), we design the NUD task to distinguish whether the translated text is coherent to be the next utterance of the given dialogue history context. Concretely, we construct positive and negative samples of context-utterance pairs from the chat corpus. A positive sample ($\mathcal{C}_{Y_{u}}$, ${Y}_{u^{+}}$) with the label $\ell=1$ consists of the target utterance ${Y}_{u}$ and its dialogue history context $\mathcal{C}_{Y_{u}}$. A negative sample ($\mathcal{C}_{Y_{u}}$, ${Y}_{u^{-}}$) with the label $\ell=0$ consists of the identical $\mathcal{C}_{Y_{u}}$ and a randomly selected utterance ${Y}_{u^{-}}$ from the training set. Formally, the training objective of NUD is defined as follows:
\begin{equation}
\begin{split}
    \mathcal{L}_{\text{NUD}}  =&-\mathrm{log}(p(\ell=1|\mathcal{C}_{Y_{u}}, {Y}_{u^{+}}))\\ &- \mathrm{log}(p(\ell=0|\mathcal{C}_{Y_{u}}, {Y}_{u^{-}})).
\end{split}
\label{loss_NUD}
\end{equation}

For a training sample ($\mathcal{C}_{Y_u}$, ${Y}_u$), to estimate the probability in \autoref{loss_NUD} for discrimination, we first obtain the representations $\mathbf{H}_{Y_u}$ of the target utterance ${Y}_u$ and $\mathbf{H}_{\mathcal{C}_{Y_u}}$ of the given dialogue history context $\mathcal{C}_{Y_u}$ using the NCT encoder. Specifically, $\mathbf{H}_{Y_u}$ is calculated as $\frac{1}{|Y_u|}\sum_{t=1}^{|Y_u|}\mathbf{h}^{L}_{e,t}$ while $\mathbf{H}_{\mathcal{C}_{Y_u}}$ is defined as the encoder hidden state $\mathbf{h}^{L}_{e,0}$ of the prepended special token `[cls]' of $\mathcal{C}_{Y_u}$. Then, the concatenation of $\mathbf{H}_{Y_u}$ and $\mathbf{H}_{\mathcal{C}_{Y_u}}$ is fed into a binary NUD classifier, which is an extra fully-connected layer on top of the NCT encoder:
\begin{equation}\nonumber
\resizebox{1.0\hsize}{!}{$
\begin{split}
    {p}(\ell \!=\!1|\mathcal{C}_{Y_u}, Y_u)\!=\!\mathrm{Softmax}(\mathbf{W}_{n}[\mathbf{H}_{Y_u}; \mathbf{H}_{\mathcal{C}_{Y_u}}]),\\
\end{split}
$}
\end{equation}
where $\mathbf{W}_{n}$ is the trainable parameter of the NUD classifier and the bias term is omitted for simplicity.

\subsubsection{Speaker Personality Modeling} 
A dialogue always involves speakers who have different personalities, which is a salient characteristic of conversations. Therefore, we design a speaker identification task that incorporates the modeling of speaker personality into the NCT model, making the translated utterance more speaker-relevant.

\paragraph{Speaker Identification (SI).}
As explored in \cite{bak-oh-2019-variational,wu-etal-2020-guiding,liang2020infusing,DBLP:conf/acl/LinYYLZLHS20}, the history utterances of a speaker can reflect a distinctive personality. \autoref{fig:subtask}(d) depicts the SI task in detail, where the NCT model is used to distinguish whether a translated utterance and a given speaker-specific history utterances are spoken by the same speaker. We also construct positive and negative training samples from the chat corpus. A positive sample ($\mathcal{C}^{sx}_{Y_{u}}$, ${Y}_{u}$) with the label $\ell=1$ consists of the target utterance ${Y}_{u}$ and the speaker $sx$-specific history context $\mathcal{C}^{sx}_{Y_{u}}$, because ${Y_{u}}$ is the translation of the utterance originally spoken by the speaker $sx$. A negative sample ($\mathcal{C}^{sy}_{Y_{u}}$, ${Y}_{u}$) with the label $\ell=0$ consists of the target utterance ${Y}_{u}$ and the speaker $sy$-specific history context $\mathcal{C}^{sy}_{Y_{u}}$. Formally, the training objective of SI is defined as follows:
\begin{equation}
\begin{split}
    \mathcal{L}_{\text{SI}}  =&-\mathrm{log}(p(\ell=1|\mathcal{C}^{sx}_{Y_{u}}, {Y}_u))\\ &- \mathrm{log}(p(\ell=0|\mathcal{C}^{sy}_{Y_{u}}, {Y}_u)).
\end{split}
\label{loss_SI}
\end{equation}

For a training sample ($\mathcal{C}^{s}_{Y_{u}}$, ${Y}_u$) with $s$$\in$$\{sx,sy\}$, we also use the NCT encoder to obtain the representations $\mathbf{H}_{Y_u}$ of the target utterance ${Y}_u$ and $\mathbf{H}_{\mathcal{C}^{s}_{Y_u}}$ of the given speaker-specific history context $\mathcal{C}^{s}_{Y_u}$. Similar to the NUD task, $\mathbf{H}_{Y_u}$=$\frac{1}{|Y_u|}\sum_{t=1}^{|Y_u|}\mathbf{h}^{L}_{e,t}$ and the $\mathbf{h}^{L}_{e,0}$ of $\mathcal{C}^{s}_{Y_u}$ is used as $\mathbf{H}_{\mathcal{C}^{s}_{Y_u}}$. Then, to estimate the probability in~\autoref{loss_SI}, the concatenation of $\mathbf{H}_{Y_u}$ and $\mathbf{H}_{\mathcal{C}^{s}_{Y_u}}$ is fed into a binary SI classifier, which is another fully-connected layer on top of the NCT encoder:
\begin{equation}\nonumber
\resizebox{1.0\hsize}{!}{$
\begin{split}
    {p}(\ell\!=\!1|\mathcal{C}^{s}_{Y_u}, Y_u)\!=\!\mathrm{Softmax}(\mathbf{W}_{s}[\mathbf{H}_{Y_u}; \mathbf{H}_{\mathcal{C}^{s}_{Y_u}}]),\\
\end{split}
$}
\end{equation}
where $\mathbf{W}_{s}$ is the trainable parameter of the SI classifier and the bias term is also omitted.

\begin{algorithm}[!t]
    \caption{Optimization Algorithm}
			\SetKwData{Index}{Index}
			\SetKwInput{kwInit}{Init}
			\SetKwInput{kwOutput}{Output}
			\SetKwInput{kwInput}{Input}
			\label{alg}
    {
		\kwInput{Sentence-level/Chat-level translation data $\mathcal{D}^s$/ $\mathcal{D}^c$,  Sentence-level/Chat-level MaxStep $T_1$/$T_2$, CoherenceMaxStep $T_2$, SpeakerMaxStep $T_2$\\}
		\kwInit{ $\theta$} 
        \Indp
        
        \Indm
        $t_1=0$    (\emph{Training sentence-level NMT model})\\
        \For{$t_1$ $<$ $T_1$}{
            Randomly sample a batch $k$ from $\mathcal{D}^s$.\\
            Compute $\mathcal{L}_{\text{S-NMT}}$.\\
            Update the parameters of the standard transformer model using Adam.\\
        }
        \kwOutput{  $\theta$}
        \kwInit{$\Theta$ using $\theta$, $\alpha=1.0$, $\beta=1.0$} 
        $t_2=0$ (\emph{Training chat-level NMT model})\\
        \For{$t_2$ $<$ $T_2$}{
            Randomly sample a batch $k$ from $\mathcal{D}^c$.\\
            Compute $\mathcal{L}_{\text{MRG}}$, $\mathcal{L}_{\text{CRG}}$, $\mathcal{L}_{\text{NUD}}$, $\mathcal{L}_{\text{SI}}$, and $\mathcal{L}_{\text{NCT}}$.\\

            Update the parameters of the CSA-NCT model with respect to $\mathcal{J}$ using Adam.\\
            $d_1=\alpha*t_2/T_2$,\quad $d_2=\beta*t_2/T_2$\\
            $\alpha = max(0,\alpha - d_1)$\\
            $\beta = max(0,\beta - d_2)$ \\
        }
        \kwOutput{  $\Theta$}
    }
\end{algorithm}

\subsection{Training and Inference}
\label{sec:ti}
For training, with the main chat translation task and four auxiliary tasks, the total training objective is finally formulated as
\begin{equation}
\begin{split}
    &\mathcal{J} = \mathcal{L}_{\text{NCT}} + \alpha(\mathcal{L}_{\text{MRG}} + \mathcal{L}_{\text{CRG}} + \mathcal{L}_{\text{NUD}}) + \beta\mathcal{L}_{\text{SI}},
\end{split}\label{loss_all}
\end{equation}
where $\alpha$ and $\beta$ are balancing hyper-parameters for the trade-off between $\mathcal{L}_{\text{NCT}}$ and the other auxiliary objectives. 
Algorithm~\ref{alg} summarizes the training procedure of the above multi-task learning process, where $\theta$ refers to the parameters of our NCT model and $\Theta$ refers to the whole set of parameters including both $\theta$ and the parameters of the additional classifiers for auxiliary tasks.

During inference, the four auxiliary tasks are not involved and only the NCT model ($\theta$) is used to conduct chat translation.

\section{Experiments}
\subsection{Datasets and Metrics}
\label{sect:data}
\paragraph{Datasets.} As shown in Algorithm~\ref{alg}, the training of our CSA-NCT framework consists of two stages: (1) pre-train the model on a large-scale sentence-level NMT corpus (WMT20\footnote{http://www.statmt.org/wmt20/translation-task.html}); (2) fine-tune on the chat translation corpus (BConTrasT~\cite{farajian-etal-2020-findings} and BMELD~\cite{liang-etal-2021-modeling}). The dataset details (\emph{e.g.}, splits of training, validation or test sets) are described in Appendix A.

\noindent \textbf{Metrics.}
For fair comparison, we use the SacreBLEU\footnote{BLEU+case.mixed+numrefs.1+smooth.exp+tok.13a+\\version.1.4.13}~\cite{post-2018-call} and TER~\cite{snover2006study} with the statistical significance test~\cite{koehn-2004-statistical}. For En$\Leftrightarrow$De, we report case-sensitive score following the WMT20 chat task~\cite{farajian-etal-2020-findings}. For Zh$\Rightarrow$En, we report case-insensitive score. For En$\Rightarrow$Zh, the reported SacreBLEU is at the character level. 

\subsection{Implementation Details}
In this paper, we adopt the settings of standard \emph{Transformer-Base} and \emph{Transformer-Big} in~\cite{vaswani2017attention} and follow the main setting in~\cite{liang-etal-2021-modeling}. Specifically, in \emph{Transformer-Base}, we use 512 as hidden size (\emph{i.e.}, $d$), 2048 as filter size and 8 heads in multihead attention. In \emph{Transformer-Big}, we use 1024 as hidden size, 4096 as filter size, and 16 heads in multihead attention. All our Transformer models contain $L$ = 6 encoder layers and $L$ = 6 decoder layers and all models are trained using THUMT~\cite{tan-etal-2020-thumt} framework. The training step for the first pre-training stage is set to $T_1$ = 200,000 while that of the second fine-tuning stage is set to $T_2$ = 5,000. The batch size for each GPU is set to 4096 tokens. All experiments in the first stage are conducted utilizing 8 NVIDIA Tesla V100 GPUs, while we use 4 GPUs for the second stage, \emph{i.e.}, fine-tuning. That gives us about 8*4096 and 4*4096 tokens per update for all experiments in the first-stage and second-stage, respectively. All models are optimized using Adam~\cite{kingma2017adam} with $\beta_1$ = 0.9 and $\beta_2$ = 0.998, and learning rate is set to 1.0 for all experiments. Label smoothing is set to 0.1. We use dropout of 0.1/0.3 for \emph{Base} and \emph{Big} setting, respectively. $|T|$ is set to 10. Following~\cite{liang-etal-2021-modeling}, we set the number of preceding sentences to 3 in all experiments.
The criterion for selecting hyper-parameters is the BLEU score on validation sets for both tasks. During inference, the beam size is set to 4, and the length penalty is 0.6 among all experiments. 

\subsection{Effect of $\alpha$ and $\beta$}
We also investigate the effect of balancing factor $\alpha$ and $\beta$, where $\alpha$ and $\beta$ gradually decrease from 1 to 0 over 5,000 steps, which is similar to~\cite{zhao-etal-2020-learning-simple}. ``Fixed $\alpha$ and $\beta$'' means we keep $\alpha$ = $\beta$ = 1 across the training. ``Dynamic $\alpha$ and $\beta$'' denotes decaying $\alpha$ and $\beta$ with the training step of auxiliary tasks. The results of \autoref{alpha_and_beta} show that ``Dynamic $\alpha$ and $\beta$'' gives better performance than ``Fixed $\alpha$ and $\beta$''. Therefore, we apply this dynamic strategy in the following experiments.

\begin{table}[t]
\centering
\newcommand{\tabincell}[2]{\begin{tabular}{@{}#1@{}}#2\end{tabular}}
\small
\begin{tabular}{l|c|c|c}
\toprule
\hline\vspace{-2pt}
&\multirow{1}{*}{Setting} & \multicolumn{1}{c|}{$\textbf{En$\Rightarrow$De}$}& \multicolumn{1}{c}{$\textbf{En$\Rightarrow$Zh}$} \\\hline
\multirow{2}{*}{\tabincell{c}{\emph{Big}}}
&Fixed $\alpha$ and $\beta$  &60.91/24.6  &29.69/55.4  \\
&Dynamic $\alpha$ and $\beta$   &61.27/24.3  &30.52/54.6 \\
\bottomrule
\end{tabular}
\caption{The BLEU/TER score (\%) results on the validation sets.
}\label{alpha_and_beta} 
\end{table}

\begin{table*}[t!]
\centering
\newcommand{\tabincell}[2]{\begin{tabular}{@{}#1@{}}#2\end{tabular}}
\setlength{\tabcolsep}{0.4mm}{
\begin{tabular}{c|l|ll|ll|ll|ll}
\toprule
\hline
\vspace{-2pt}
&\multicolumn{1}{c|}{\multirow{2}{*}{\textbf{Models}}} &\multicolumn{2}{c|}{$\textbf{En$\Rightarrow$De}$}  &  \multicolumn{2}{c|}{$\textbf{De$\Rightarrow$En}$}    &\multicolumn{2}{c|}{$\textbf{En$\Rightarrow$Zh}$}  &  \multicolumn{2}{c}{$\textbf{Zh$\Rightarrow$En}$} \\ 
&\multicolumn{1}{c|}{} & \multicolumn{1}{c}{BLEU$\uparrow$} & \multicolumn{1}{c|}{TER$\downarrow$} & \multicolumn{1}{c}{BLEU$\uparrow$} &  \multicolumn{1}{c|}{TER$\downarrow$}      & \multicolumn{1}{c}{BLEU$\uparrow$} & \multicolumn{1}{c|}{TER$\downarrow$} & \multicolumn{1}{c}{BLEU$\uparrow$} & \multicolumn{1}{c}{TER$\downarrow$}   \\ \hline\hline
\multirow{2}{*}{\tabincell{c}{\emph{Sentence-Level}\\\emph{NMT models (Base)}}}
&{Transformer}    & 40.02     & 42.5   & 48.38     & 33.4      &21.40     &72.4 & 18.52    & 59.1  \\
&{Transformer+FT}   & 58.43 & 26.7   & \underline{59.57}& 26.2 &25.22 &62.8 & \underline{21.59} &\underline{56.7}\\\cdashline{1-10}[4pt/2pt]
\multirow{4}{*}{\tabincell{c}{\emph{Context-Aware}\\\emph{NMT models (Base)}}}
&{Dia-Transformer+FT}  &58.33     &26.8   &59.09    &26.2   &24.96     &63.7 & 20.49   & 60.1   \\
&{Doc-Transformer+FT}  & 58.15     & 27.1   &59.46    &\underline{25.7}  & 24.76     &63.4  & 20.61   & 59.8   \\ 
&{Gate-Transformer+FT}  &\underline{58.48}     &\underline{26.6}   &59.53    &26.1   &\underline{25.34}     &\underline{62.5} &21.03   &56.9   \\\cdashline{2-10}[4pt/2pt]
&{CSA-NCT (Ours)}   & \textbf{59.50}$^{\dagger\dagger}$  & \textbf{25.7}$^{\dagger\dagger}$  &\textbf{60.65}$^{\dagger\dagger}$   &\textbf{25.4}$^{\dagger}$  &\textbf{27.77}$^{\dagger\dagger}$     &\textbf{60.0}$^{\dagger\dagger}$  & \textbf{22.36}$^{\dagger}$   & \textbf{55.9}$^{\dagger}$   \\\hline\hline
\multirow{2}{*}{\tabincell{c}{\emph{Sentence-Level}\\\emph{NMT models (Big)}}}
&{Transformer}    &40.53     &42.2   &49.90 &33.3   &22.81  &69.6     &19.58    &57.7 \\
&{Transformer+FT}   &\underline{59.01} &\underline{26.0} & 59.98  &25.9   &26.95 &60.7 &22.15 &56.1\\\cdashline{1-10}[4pt/2pt]
\multirow{4}{*}{\tabincell{c}{\emph{Context-Aware}\\\emph{NMT models (Big)}}}
&{Dia-Transformer+FT}  &58.68    &26.8  &59.63 &26.0   &26.72  &62.4    &21.09    &58.1  \\ 
&{Doc-Transformer+FT}  &58.61    &26.5 &59.98 &\underline{25.4}   &26.45  &62.6   &21.38    &57.7  \\ 
&{Gate-Transformer+FT}  &58.94     &26.2   &\underline{60.08}    &25.5   &\underline{27.13}     &\underline{60.3} &\underline{22.26}   &\underline{55.8}   \\\cdashline{2-10}[4pt/2pt]
&{CSA-NCT (Ours)}   &\textbf{60.64}$^{\dagger\dagger}$  &\textbf{25.3}$^{\dagger}$ &\textbf{61.21}$^{\dagger\dagger}$  &\textbf{24.9}$^{\dagger}$ &\textbf{28.86}$^{\dagger\dagger}$  &\textbf{58.7}$^{\dagger\dagger}$  &\textbf{23.69}$^{\dagger\dagger}$  &\textbf{54.7}$^{\dagger\dagger}$\\ 
\bottomrule
\end{tabular}}
\caption{Results on the test sets of BConTrasT (En$\Leftrightarrow$De) and BMELD (En$\Leftrightarrow$Zh) in terms of BLEU (\%) and TER (\%). The best and the second results are bold and underlined, respectively. ``$^{\dagger}$'' and ``$^{\dagger\dagger}$'' indicate that statistically significant better than the best result of all contrast NMT models with t-test {\em p} \textless \ 0.05 and {\em p} \textless \ 0.01, respectively. All ``+FT'' models apply the same two-stage training strategy with our CSA-NCT model for fair comparison.}
\label{tbl:main_res}
\end{table*}

\subsection{Comparison Models}
\label{ssec:layout}
\textbf{Baseline Sentence-Level NMT Models.}
\begin{itemize}
\item \textbf{Transformer}~\cite{vaswani2017attention}: The de-facto NMT model trained on sentence-level NMT corpus. 
\item \textbf{Transformer+FT}~\cite{vaswani2017attention}: The NMT model that is directly fine-tuned on the chat translation data after being pre-trained on sentence-level NMT corpus.
\end{itemize}
\noindent\textbf{Existing Context-Aware NMT Systems.}
\begin{itemize}
\item \textbf{Dia-Transformer+FT}~\cite{maruf-etal-2018-contextual}: The original model is RNN-based and an additional encoder is used to incorporate the mixed-language dialogue history. We re-implement it based on Transformer where an additional encoder layer is used to introduce the dialogue history into NMT model.
\item \textbf{Doc-Transformer+FT}~\cite{ma-etal-2020-simple}: A state-of-the-art document-level NMT model based on Transformer sharing the first encoder layer to incorporate the dialogue history.
\item \textbf{Gate-Transformer+FT}~\cite{zhang-etal-2018-improving}: A document-aware Transformer that uses a gate to incorporate the context information. Note that we share the Transformer encoder to obtain the context representation instead of utilizing the additional context encoder, which performs better in our experiments.
\end{itemize}


\subsection{Main Results}
In \autoref{tbl:main_res}, We report the main results on En$\Leftrightarrow$De and En$\Leftrightarrow$Zh under \emph{Base} and \emph{Big} settings. For comparison, as in \autoref{ssec:layout}, ``Transformer'' and ``Transformer+FT'' are sentence-level baselines while ``Dia-Transformer+FT'', ``Doc-Transformer+FT'' and ``Gate-Transformer+FT'' are the existing context-aware NMT systems re-implemented by us. Particularly, ``CSA-NCT'' represents our proposed approach.

\paragraph{Results on En$\Leftrightarrow$De.}
\label{ssec:ende}
Under the \emph{Base} setting, our model substantially outperforms the sentence-level/context-aware baselines by a large margin (\emph{e.g.}, the previous best ``Gate-Transformer+FT''), 1.02$\uparrow$ on En$\Rightarrow$De and 1.12$\uparrow$ on De$\Rightarrow$En. In term of TER, CSA-NCT also performs better on the two directions, 0.9$\downarrow$ and 0.7$\downarrow$ lower than ``Gate-Transformer+FT'' (the lower the better), respectively. Under the \emph{Big} setting, on En$\Rightarrow$De and De$\Rightarrow$En, our model consistently surpasses the baselines and other existing systems again.

\paragraph{Results on En$\Leftrightarrow$Zh.}
\label{ssec:chen}
We also conduct experiments on the BMELD dataset. Concretely, on En$\Rightarrow$Zh and Zh$\Rightarrow$En, our model also presents notable improvements over all comparison models by at least 2.43$\uparrow$ and 0.77$\uparrow$ BLEU gains under the \emph{Base} setting, and by 1.73$\uparrow$ and 1.43$\uparrow$ BLEU gains under the \emph{Big} setting, respectively. These results demonstrate the effectiveness and generalizability of our model across different language pairs.

\begin{table}[t!]
\centering
\scalebox{0.86}{
\setlength{\tabcolsep}{0.30mm}{
\begin{tabular}{l|l|cc|cc}
\toprule
\hline
\vspace{-2pt}
\multirow{2}{*}{\#}&\multicolumn{1}{c|}{\multirow{2}{*}{\textbf{Models}}} &\multicolumn{2}{c|}{$\textbf{En$\Rightarrow$De}$}  &  \multicolumn{2}{c}{$\textbf{De$\Rightarrow$En}$}    \\ 
&\multicolumn{1}{c|}{} & \multicolumn{1}{c}{BLEU$\uparrow$} & \multicolumn{1}{c|}{TER$\downarrow$} & \multicolumn{1}{c}{BLEU$\uparrow$} & \multicolumn{1}{c}{TER$\downarrow$}   \\ \hline
0&{Baseline}   & {60.40}  & {25.0} &{61.68}    & {24.9} \\\cdashline{1-6}[4pt/2pt] 
1&\emph{w}/ {DCM } & 61.05$^{\dagger\dagger}$ (+0.65)  & 24.4$^{\dagger\dagger}$  & 62.63$^{\dagger\dagger}$ (+0.95) & 24.5$^{\dagger}$      \\ 
2&\emph{w}/  SPM & 60.57 (+0.17)  & 24.8    &61.97 (+0.29)    &24.7  \\
\bottomrule
\end{tabular}}}
\caption{Ablation results on the validation sets of each auxiliary task group under the \emph{Big} setting. ``Baseline'' represents the NCT model without any auxiliary task. ``DCM'': dialogue coherence modeling, including MRG, CRG, NUD. ``SPM'': speaker personality modeling, \emph{i.e.}, SI. ``$^{\dagger}$'' and ``$^{\dagger\dagger}$'' indicate the improvement over the result of the baseline model is statistically significant with {\em p} \textless \ 0.05 and {\em p} \textless \ 0.01), respectively.}
\label{tbl:ablation}
\end{table}
\section{Analysis}
\subsection{Ablation Study}
\label{ssec:abs}
\paragraph{Effect of Each Auxiliary Task Group.} We conduct ablation studies to investigate the effects of the two groups (DCM and SPM) of auxiliary tasks. The results under the \emph{Big} setting are listed in \autoref{tbl:ablation}. We have the following findings: (1) DCM substantially improves the NCT model in terms of both BLEU and TER metrics, which demonstrates modeling coherence is beneficial for better translations.
(2) SPM makes slight contributions to the NCT model in terms of BLEU, which is less significant than DCM. However, further human evaluation in~\autoref{ssec:he} will show that our model can keep the personality consistent with the original speaker.  

\begin{table}[t!]
\centering
\scalebox{0.86}{
\setlength{\tabcolsep}{0.30mm}{
\begin{tabular}{l|l|cc|cc}
\toprule
\hline
\vspace{-2pt}
\multirow{2}{*}{\#}&\multicolumn{1}{c|}{\multirow{2}{*}{\textbf{Models}}} &\multicolumn{2}{c|}{$\textbf{En$\Rightarrow$De}$}  &  \multicolumn{2}{c}{$\textbf{De$\Rightarrow$En}$}    \\ 
&\multicolumn{1}{c|}{} & \multicolumn{1}{c}{BLEU$\uparrow$} & \multicolumn{1}{c|}{TER$\downarrow$} & \multicolumn{1}{c}{BLEU$\uparrow$} & \multicolumn{1}{c}{TER$\downarrow$}   \\ \hline
0&Baseline & {60.40}  & {25.0} &{61.68}    & {24.9} \\\cdashline{1-6}[4pt/2pt]
1&\emph{w}/ {MRG } & 61.00$^{\dagger\dagger}$ (+0.60)   & 24.4$^{\dagger\dagger}$    & 62.37$^{\dagger\dagger}$ (+0.69) & 24.5$^{\dagger}$      \\ 
2&\emph{w}/ {CRG}  & 60.68 (+0.28)  & 24.6$^{\dagger}$    &62.14$^{\dagger}$ (+0.46)   & 24.8  \\ 
3&\emph{w}/ {NUD} & 60.82$^{\dagger}$ (+0.42)  & 24.7  & 62.32$^{\dagger\dagger}$ (+0.64) & 24.7  \\
4&\emph{w}/  SI & 60.57 (+0.17)  & 24.8    &61.97 (+0.29)    &24.7  \\
\bottomrule
\end{tabular}}}
\caption{Ablation results on the validation sets of each auxiliary task under the \emph{Big} setting. ``$^{\dagger}$'' and ``$^{\dagger\dagger}$'' indicate the improvement over the result of the baseline model is statistically significant with {\em p} \textless \ 0.05 and {\em p} \textless \ 0.01, respectively.}
\label{tbl:ablation2}
\end{table}

\paragraph{Effect of Each Auxiliary Task.}
We also investigate the effect of each auxiliary task by adding a single task at a time. In \autoref{tbl:ablation2}, rows 1$\sim$4 denote singly adding on the corresponding auxiliary task with the main chat translation task, each of which shows a positive impact on the model performance (rows 1$\sim$4 vs. row 0). 
\subsection{Dialogue Coherence}
Following~\cite{lapata2005automatic,Xiong_He_Wu_Wang_2019}, we measure dialogue coherence as sentence similarity, which is determined by the cosine similarity between two sentences $s_1$ and $s_2$: 
\begin{equation}\nonumber
\begin{split}
    sim(s_1, s_2) &= \mathrm{cos}(f({s_1}), f({s_2})),
\end{split}
\end{equation}
where $f(s_i) = \frac{1}{\vert s_i\vert}\sum_{\textbf{w} \in s_i}(\textbf{w})$ and \(\textbf{w}\) is the vector for word $w$. We use Word2Vec\footnote{https://code.google.com/archive/p/word2vec/}~\cite{mikolov2013efficient} trained on a dialogue dataset\footnote{Due to no available German dialogue datasets, we choose Taskmaster-1~\cite{byrne-etal-2019-taskmaster}, where the English side of BConTrasT~\cite{farajian-etal-2020-findings} also comes from it.} to obtain the distributed word vectors whose dimension is set to 100.

\autoref{coherence} shows the measured coherence of different models on the test set of BConTrasT in De$\Rightarrow$En direction. It shows that our CSA-NCT produces more coherent translations compared to baselines and other existing systems (significance test, {\em p} \textless \ 0.01).
\label{ssec:dc}
\begin{table}[t]
\centering
\newcommand{\tabincell}[2]{\begin{tabular}{@{}#1@{}}#2\end{tabular}}
\setlength{\tabcolsep}{0.6mm}{
\begin{tabular}{l|c|c|c}
\toprule
\hline
\vspace{-2pt}
\multirow{1}{*}{\textbf{Models}} &  \multicolumn{1}{c|}{\textbf{1-th Pr.}} & \multicolumn{1}{c|}{\textbf{2-th Pr.}}  & \multicolumn{1}{c}{\textbf{3-th Pr.}}\\\cline{1-4}
Transformer               &65.02  &60.37 &56.59\\
Transformer+FT            &65.87  &61.04 &57.14  \\\cdashline{1-4}[4pt/2pt]
Dia-Transformer+FT        &65.53  &60.84 &57.09\\
Doc-Transformer+FT        &65.69  &60.93 &57.13\\
{Gate-Transformer+FT}     &65.96  &61.35 &57.45\\\cdashline{2-4}[4pt/2pt]
CSA-NCT (Ours)  &66.57$^{\dagger\dagger}$  &{61.78}$^{\dagger\dagger}$ &{57.83}$^{\dagger\dagger}$  \\\cdashline{1-4}[4pt/2pt]
Human Reference         &\textbf{66.63}  &\textbf{61.90} &\textbf{57.95}\\
\bottomrule
\end{tabular}}
\caption{Results (\%) of dialogue coherence in terms of sentence similarity on the test set of BConTrasT in De$\Rightarrow$En direction under the \emph{Base} setting. The ``\#\textbf{-th Pr.}'' denotes the \#-th preceding utterance to the current one. ``$^{\dagger\dagger}$'' indicates the improvement over the best result of all other comparison models is statistically significant ({\em p} \textless \ 0.01).}
\label{coherence} 
\end{table}

\begin{table}[t]
\centering
\newcommand{\tabincell}[2]{\begin{tabular}{@{}#1@{}}#2\end{tabular}}
\setlength{\tabcolsep}{1.8mm}{
\begin{tabular}{l|c|c|c}
\toprule
\hline
\vspace{-2pt}
\multirow{1}{*}{\textbf{Models}} & \multicolumn{1}{c|}{$\textbf{Coh.}$} &  \multicolumn{1}{c|}{$\textbf{Spe.}$} &  \multicolumn{1}{c}{$\textbf{Flu.}$} \\\cline{1-4}
Transformer        &0.540 &0.485  &0.590 \\
Transformer+FT     &0.590 &0.530  &0.635 \\\cdashline{1-4}[4pt/2pt]
Dia-Transformer+FT &0.580   &0.525  &0.625 \\
Doc-Transformer+FT &0.595   &0.525  &0.630 \\
{Gate-Transformer+FT}  &0.605     &0.540   &0.635 \\\cdashline{2-4}[4pt/2pt]
CSA-NCT (Ours)       &\textbf{0.635} &\textbf{0.575}  &\textbf{0.655} \\
\bottomrule
\end{tabular}}
\caption{Results of Human evaluation ({Zh$\Rightarrow$En}, \emph{Base}).  ``\textbf{Coh.}'': Coherence. ``\textbf{Spe.}'': Speaker. ``\textbf{Flu.}'': Fluency. }
\label{human_evaluation} 
\end{table}

\subsection{Human Evaluation}
\label{ssec:he}
Inspired by~\cite{bao-EtAl:2020:WMT,farajian-etal-2020-findings}, we use three criteria for human evaluation: (1) \textbf{Coherence} measures whether the translation is semantically coherent with the dialogue history; (2) \textbf{Speaker} measures whether the translation preserves the personality of the speaker; (3) \textbf{Fluency} measures whether the translation is fluent and grammatically correct.

First, we randomly sample 200 conversations from the test set of BMELD in Zh$\Rightarrow$En direction. Then, we use the 6 models in \autoref{human_evaluation} to generate the translated utterances of these sampled conversations. Finally, we assign the translated utterances and their corresponding dialogue history utterances in target language to three postgraduate human annotators, and ask them to make evaluations from the above three criteria.

The results in \autoref{human_evaluation} show that our model generates more coherent, speaker-relevant, and fluent translations compared with other models (significance test, {\em p} \textless \ 0.05), indicating the superiority of our model. The inter-annotator agreements calculated by the Fleiss’ kappa~\cite{doi:10.1177/001316447303300309} are 0.506, 0.548, and 0.497 for {coherence}, {speaker} and {fluency}, respectively, indicating ``Moderate Agreement'' for all four criteria. We also present one case study in Appendix B.

\label{ssec:cs}

\section{Related Work}
\paragraph{Chat NMT.} Little prior work is available due to the lack of human-annotated publicly available data~\cite{farajian-etal-2020-findings}. Therefore, some existing studies~\cite{lrec,maruf-etal-2018-contextual,9023129,rikters-etal-2020-document} mainly pay attention to designing methods to automatically construct the subtitle corpus, which may contain noisy bilingual utterances. Recently, ~\citet{farajian-etal-2020-findings} organize the WMT20 chat translation task and first provide a chat corpus post-edited by humans. More recently, based on document-level parallel corpus, ~\citet{wang2021autocorrect} propose to jointly identify omissions and typos within dialogue along with translating utterances by using the context. As a concurrent work, ~\citet{liang-etal-2021-modeling} provide a clean bilingual dialogue dataset and design a variational framework for NCT. Different from them, we focus on introducing the modeling of dialogue coherence and speaker personality into the NCT model with multi-task learning to promote the translation quality. 

\paragraph{Context-Aware NMT.} In a sense, chat MT can be viewed as a special case of context-aware MT that has many related studies~\cite{gong-etal-2011-cache,jean2017does,wang-etal-2017-exploiting-cross,zheng2020making,yang-etal-2019-enhancing,kang-etal-2020-dynamic,li-etal-2020-multi,DBLP:journals/corr/abs-2006-04721,ma-etal-2020-simple}. Typically, they resort to extending conventional NMT models for exploiting the context. Although these models can be directly applied to the chat translation scenario, they cannot explicitly capture the inherent dialogue characteristics and usually lead to incoherent and speaker-irrelevant translations. 

\section{Conclusion}
In this paper, we propose to enhance the NCT model by introducing the modeling of the inherent dialogue characteristics, \emph{i.e.}, dialogue coherence and speaker personality. We train the NCT model with the four well-designed auxiliary tasks, \emph{i.e.}, MRG, CRG, NUD and SI. Experiments on En$\Leftrightarrow$De and En$\Leftrightarrow$Zh show that our model notably improves translation quality on both BLEU and TER metrics, showing its superiority and generalizability. Human evaluation further verifies that our model yields more coherent and speaker-relevant translations. 

\section*{Acknowledgements}
The research work descried in this paper has been supported by the National Key R\&D Program of China (2020AAA0108001) and the National Nature Science Foundation of China (No. 61976015, 61976016, 61876198 and 61370130). The authors would like to thank the anonymous reviewers for their valuable comments and suggestions to improve this paper.

\bibliography{anthology,custom}
\bibliographystyle{acl_natbib}

\appendix
\label{sec:appendix}

\section{Datasets}
As mentioned in \autoref{sect:data}, our experiments involve the dataset WMT20 for pre-training and two chat translation corpus, BConTrasT~\cite{farajian-etal-2020-findings} and BMELD~\cite{liang-etal-2021-modeling}. The statistics about the splits of training, validation, and test sets are shown in \autoref{datasets}.

\paragraph{WMT20.} Following~\cite{liang-etal-2021-modeling}, for En$\Leftrightarrow$De, we combine six corpora including Euporal, ParaCrawl, CommonCrawl, TildeRapid, NewsCommentary, and WikiMatrix. For En$\Leftrightarrow$Zh, we combine News Commentary v15, Wiki Titles v2, UN Parallel Corpus V1.0, CCMT Corpus, and WikiMatrix. First, we filter out duplicate sentence pairs and remove those whose length exceeds 80. To pre-process the raw data, we employ a series of open-source/in-house scripts, including full-/half-width conversion, unicode conversation, punctuation normalization, and tokenization~\cite{wang-EtAl:2020:WMT1}. After filtering, we apply BPE~\cite{sennrich-etal-2016-neural} with 32K merge operations to obtain subwords. Finally, we obtain 45,541,367 sentence pairs for En$\Leftrightarrow$De and 22,244,006 sentence pairs for En$\Leftrightarrow$Zh, respectively. 

We test the model performance of the first stage on \emph{newstest2019}. The results are shown in \autoref{bleu_on_one_stage}.

\begin{table}[t]
\centering
\newcommand{\tabincell}[2]{\begin{tabular}{@{}#1@{}}#2\end{tabular}}
\small
\setlength{\tabcolsep}{1.8mm}{
\begin{tabular}{l|rrr|rrr}
\toprule
\multirow{2}{*}{Datasets} & \multicolumn{3}{c|}{\# Dialogues} &  \multicolumn{3}{c}{\# Utterances} \\
&Train &Valid & Test&Train &Valid & Test\\\hline
En$\Rightarrow$De    &550&78&78 &7,629 &1,040 &1,133\\
De$\Rightarrow$En    &550&78&78  &6,216 &862 &967\\
En$\Rightarrow$Zh    &1,036&108&274  &5,560&567&1,466 \\
Zh$\Rightarrow$En    &1,036&108&274   &4,427&517&1,135 \\
\bottomrule
\end{tabular}}
\caption{Statistics of chat translation data.
}\label{datasets}
\end{table}

\begin{table}[t]
\centering
\newcommand{\tabincell}[2]{\begin{tabular}{@{}#1@{}}#2\end{tabular}}
\small
\setlength{\tabcolsep}{0.8mm}{
\begin{tabular}{l|c|c|c|c}
\toprule
\multirow{1}{*}{Models} & \multicolumn{1}{c|}{$\textbf{En$\Rightarrow$De}$} &  \multicolumn{1}{c|}{$\textbf{De$\Rightarrow$En}$} & \multicolumn{1}{c|}{$\textbf{En$\Rightarrow$Zh}$} &  \multicolumn{1}{c}{$\textbf{Zh$\Rightarrow$En}$} \\\hline

Transformer (Base)  &39.88  &40.72  &32.55  &24.42\\\hline
Transformer (Big)   &41.35  &41.56 &33.85  &24.86\\

\bottomrule
\end{tabular}}
\caption{The BLEU scores on the \emph{newstest2019} of the first stage. 
}\label{bleu_on_one_stage} 
\end{table}

\paragraph{BConTrasT.} The dataset\footnote{https://github.com/Unbabel/BConTrasT} is first provided by WMT 2020 Chat Translation Task~\cite{farajian-etal-2020-findings}, which is translated from English into German and is based on the monolingual Taskmaster-1 corpus~\cite{byrne-etal-2019-taskmaster}. The conversations (originally in English) were first automatically translated into German and then manually post-edited by Unbabel editors\footnote{www.unbabel.com} who are native German speakers. Having the conversations in both languages allows us to simulate bilingual conversations in which one speaker (customer), speaks in German and the other speaker (agent), responds in English.

\paragraph{BMELD.} The dataset is a recently released English$\Leftrightarrow$Chinese bilingual dialogue dataset, provided by~\citet{liang-etal-2021-modeling}. Based on the dialogue dataset in the MELD (originally in English)~\cite{poria-etal-2019-meld}\footnote{The MELD is a multimodal emotionLines dialogue dataset, each utterance of which corresponds to a video, voice, and text, and is annotated with detailed emotion and sentiment.}, they firstly crawled the corresponding Chinese translations from \url{https://www.zimutiantang.com/} and then manually post-edited them according to the dialogue history by native Chinese speakers who are post-graduate students majoring in English. Finally, following~\cite{farajian-etal-2020-findings}, they assume 50\% speakers as Chinese speakers to keep data balance for Zh$\Rightarrow$En translations and build the \underline{b}ilingual MELD (BMELD). For the Chinese, we follow them to segment the sentence using Stanford CoreNLP toolkit\footnote{https://stanfordnlp.github.io/CoreNLP/index.html}.

\begin{figure}[t]
    \centering
    \includegraphics[width=0.48\textwidth]{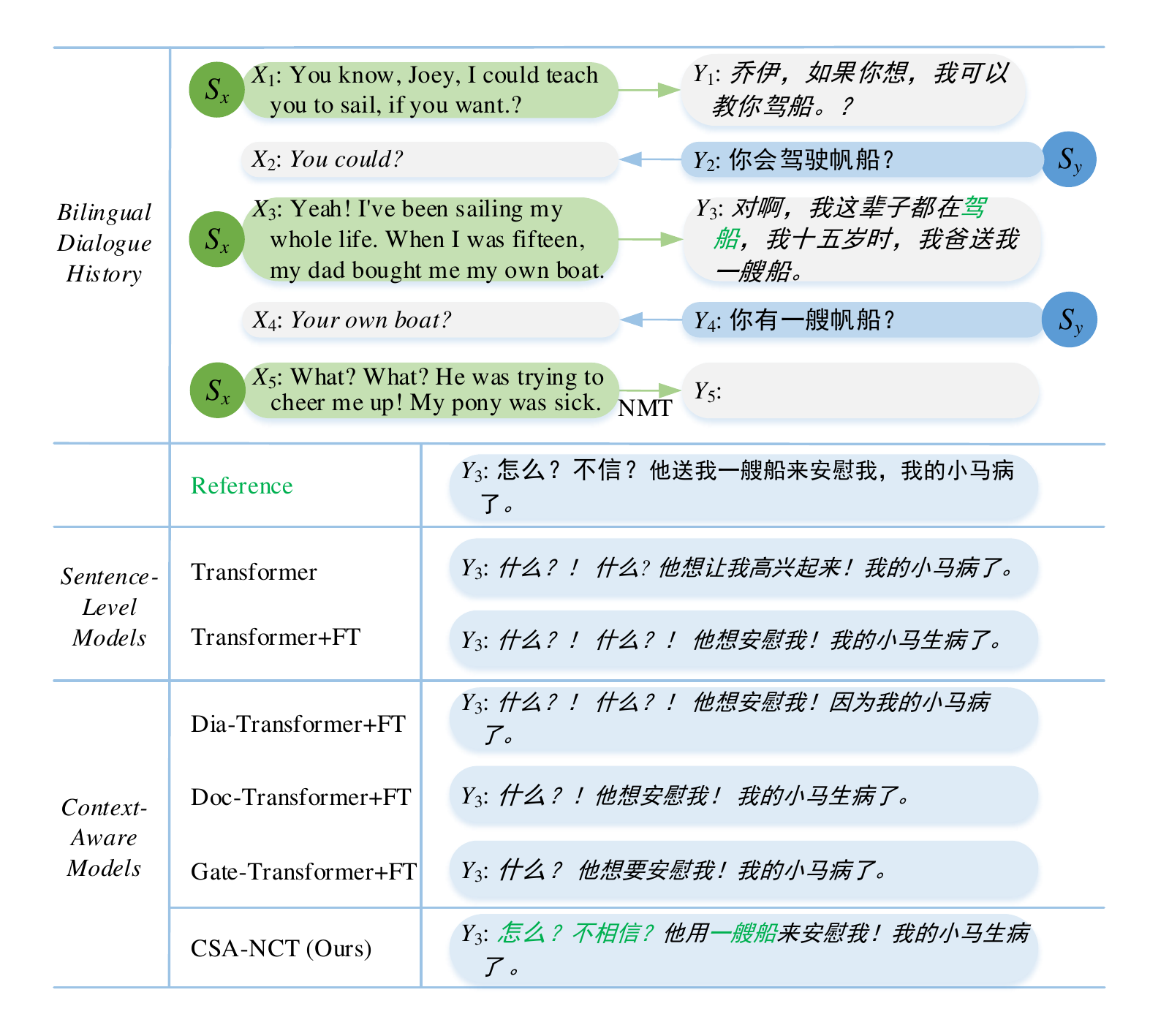}
    \caption{An illustrative case of bilingual conversation.
    }
    \label{fig:k2}
\end{figure}

\section{Case Study}

In this section, we deliver an illustrative case in \autoref{fig:k2} to show different outputs among the comparison models and ours. 
\paragraph{Dialogue Coherence and Speaker Personality.}
For the case in \autoref{fig:k2}, we find that all comparison models cannot generate coherent translated utterences. The reason may be that they fail to capture contextual clues, \emph{i.e.}, ``\emph{boat}''. By contrast, we explicitly introduce the modeling of preceding context through auxiliary tasks and thus obtain satisfactory results.  Meanwhile, we observe that the sentence-level models and the context-aware models cannot preserve the speaker personality information, \emph{e.g.}, \emph{joy} emotion, even though context-aware models incorporate the bilingual conversational history into the encoder. 

The case shows that our CSA-NCT model enhanced by the four auxiliary tasks yields coherent and speaker-relevant translations, demonstrating its effectiveness and superiority.

\end{document}